\documentclass[11pt,a4paper]{article}
\usepackage[hyperref]{emnlp2018}
\usepackage{times}
\usepackage{latexsym}

\usepackage{url}
\usepackage{graphicx}
\usepackage{enumitem}
\usepackage{multirow}
\usepackage{array}
\usepackage{todonotes}
\usepackage{subcaption}
\usepackage{hyperref}

\aclfinalcopy 

\newcommand*\samethanks[1][\value{footnote}]{\footnotemark[#1]}

\title{Training Deeper Neural Machine Translation Models \\with Transparent Attention
}

\author{Ankur Bapna \thanks{~~~Equal contribution.} \\\And
  Mia Xu Chen \samethanks[1] \\\And
  Orhan Firat \samethanks[1] \\
  {\tt ankurbpn,miachen,orhanf,yuancao@google.com}\\
  Google AI \\\And
  Yuan Cao \samethanks[1] \\\And
  Yonghui Wu \\}

\date{}

\begin{document}
\maketitle
\begin{abstract}
While current state-of-the-art NMT models, such as RNN seq2seq and Transformers, possess a large number of parameters, they are still shallow in comparison to convolutional models used for both text and vision applications. In this work we attempt to train significantly (2-3x) deeper Transformer and Bi-RNN encoders for machine translation.  We propose a simple modification to the attention mechanism that eases the optimization of deeper models, and results in consistent gains of 0.7-1.1 BLEU on the benchmark WMT'14 English-German and WMT'15 Czech-English tasks for both architectures.
\end{abstract}

\section{Introduction}

The past few years have seen significant advances in the quality of machine translation systems, owing to the advent of neural sequence to sequence models. While current state of the art models come in different flavours, including Transformers \cite{transformer}, convolutional seq2seq models \cite{convseq2seq} and LSTMs \cite{chen2018best}, all of these models follow the seq2seq with attention \cite{BahdanauCB15} paradigm.

While revolutionary new architectures have contributed significantly to these quality improvements, the importance of larger model capacities cannot be downplayed. The first major improvement in NMT quality since the switch to neural models, amongst other factors, was brought about by a huge scale up in model capacity \cite{zhou2016deep,GNMT}. While there are multiple approaches to increase capacity, deeper models have been shown to extract more expressive features \cite{Deep1,Deep2,Deep3}, and have resulted in significant gains for vision tasks over the past few years \cite{ResNet,Highway}.

Despite this being an obvious avenue for improvement, research in deeper models is often restricted by computational constraints. Additionally, deep models are often plagued by trainability concerns like vanishing or exploding gradients \cite{bengio1994learning}. These issues have been studied in the context of capturing long range dependencies in recurrent architectures \cite{DifficultyRNN,GradientFlow}, but resolving these deficiencies in Transformers or LSTM seq2seq models deeper than 8 layers is unfortunately under-explored \cite{wang2017deep,barone2017deep,devlin2017sharp}.

In this study we take the first step towards training extremely deep models for translation, by training deep encoders for Transformer and LSTM based models. As we increase the encoder depth the vanilla Transformer models completely fail to train. We also observe sub-optimal performance for LSTM models, which we believe is associated with trainability issues. To ease optimization we propose an enhancement to the attention mechanism, which allows us to train deeper models and results in consistent gains on the WMT'14 En$\rightarrow$De and WMT'15 Cs$\rightarrow$En tasks.

\section{Transparent Attention}
While the effect of attention on the forward pass is exalted with visualizations and linguistic interpretations, its influence on the gradient flow is often forgotten. Consider the original seq2seq model \textit{without} attention \cite{sutskever2014sequence}. To propagate the error signal from the last layer of the decoder to the first layer of the encoder, it has to pass through multiple time-steps in the decoder, survive the encoder-decoder bottleneck, and pass through multiple time-steps in the encoder, before reaching the parameter to be updated. There is some loss of information at every step, especially in the early stages of training. Attention \cite{BahdanauCB15} creates a direct path from the decoder to the topmost layer of the encoder, ensuring its efficient dispersal over time. This increase in inter-connectivity significantly shortens the credit-assignment path \cite{britz}, making the network less susceptible to optimization pathologies like vanishing gradients.

\begin{figure}
\centering
\includegraphics[width=\linewidth]{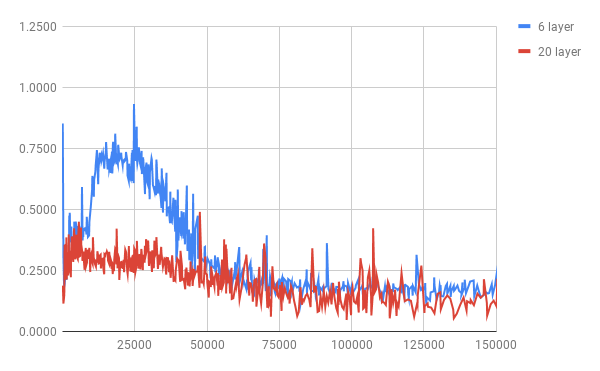}
\caption{Grad-norm ratio ($r_t$) vs training step ($t$) comparison for a 6 layer (blue) and 20 layer (red) Transformer trained on WMT 14 En$\rightarrow$De.}
\label{fig:trans_grad_norm}
\end{figure}
For deeper networks the error signal also needs to traverse along the depth of the encoder. We propose an extension to the attention mechanism that behaves akin to creating weighted residual connections along the encoder depth, allowing the dispersal of error signal simultaneously over encoder depth and time. Using trainable weights, this `transparent' attention allows the model the flexibility to adjust the gradient flow to different layers in the encoder depending on its training phase.
\vspace{-5px}
\subsection{Experimental Setup}
We train our models on the standard WMT'14 En$\rightarrow$De dataset. Each sentence is tokenized with the Moses tokenizer before breaking into sub-word units similar to 
\cite{SennrichHB15}. We use a shared vocabulary of 32k units for each
language pair. We report all our results on newstest 2014, and use a combination of newstest 2012 and newstest 2013 for validation. To verify our results, we also evaluate our models on WMT'15 Cs$\rightarrow$En. Here we use newstest 2013 for validation and newstest 2015 as the test set.
To evaluate the models we compute BLEU on the tokenized, true-case output. We report the mean post-convergence score over a window of 21 checkpoints, obtained using dev performance, following \cite{chen2018best}.
\subsection{Baseline Experiments}
We base our study on two architectures: Transformer \cite{transformer} and RNMT+ \cite{chen2018best}. We choose a smaller version of each model to fit deep encoders with up to 20 layers on a single GPU. All our models are trained on eight P100 GPUs with synchronous training, and optimized using Adam \cite{Adam}. For both architectures we train four models, with 6, 12, 16 and 20 encoder layers. We use 6 and 8 decoder layers for all our transformers and RNMT+ experiments respectively.
We also report performance for the standard Transformer Big and RNMT+ setups, as described in \cite{chen2018best}, for comparison against higher capacity models.

\textbf{Transformer}: We use the latest version of the Transformer base model, using the implementation from \cite{chen2018best}. We modify the learning rate schedule to use a learning rate of $3.0$ and $40,000$ warmup steps.

\textbf{RNMT+}: We implemented a smaller version of the En$\rightarrow$De RNMT+ model based on the description in \cite{chen2018best}, with 512 LSTM nodes in both encoder and decoder.

\begin{figure}
\centering
\includegraphics[width=\linewidth]{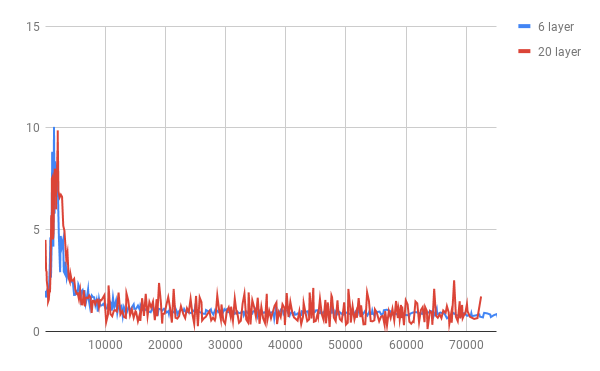}
\caption{Grad-norm ratio ($r_t$) vs training step ($t$) comparison for a 6 layer (blue) and 20 layer (red) RNMT+ model trained on WMT 14 En$\rightarrow$De.}
\label{fig:bnmt_grad_norm}
\end{figure}

\subsection{Analysis}
From Tables~\ref{tab:En-De-trans} and~\ref{tab:Cs-En-trans}, we notice that the deeper Transformer encoders completely fail to train. To understand what goes wrong we keep track of the grad norm ratio $r_t=\left(\|\nabla_{h_1} L^{(t)}\|\Big/\|\nabla_{h_N} L^{(t)}\|\right), t=1\ldots T$, where $L^{(t)}$ is the loss at time step $t$, $N$ is the number of layers in the encoder, $h_1$ is the output of the first encoder layer, $h_N$ is the output of the $N$-th encoder layer, and $T$ is the total number of training steps. We use $r_t$ as a diagnostic measure for two reasons: First, it indicates if training is suffering from exploding or vanishing gradients. Second, when a network is properly trained the lowest layers usually converge quickly, whereas the top-most layers take longer \cite{raghu2017svcca}. We therefore expect that, for a healthy training process, $r_t$ is relatively large during the early stages of training when updates to lower layers are larger than upper layers. We observe this in most successful Transformer and RNMT+ training runs.

Figure~\ref{fig:trans_grad_norm} illustrates the $r_t$ curves for the 6-layer and 20-layer Transformers. As expected, the shallow model has a high $r_t$ value during early stages of training. For the deep model, however, $r_t$ remains flat at a much smaller value throughout training. We also observe that $r_t$ remains below $1.0$ for both models, although the problem seems much less severe for the shallow model.

From Tables~\ref{tab:En-De-rnmt} and~\ref{tab:Cs-En-rnmt}, we also observe that the performance of deep RNMT+ encoders is not significantly impacted, reaching the level of the 6 layer model. This is supported by the RNMT+ $r_t$ curves in Figure \ref{fig:bnmt_grad_norm}, which indicate few differences in the learning dynamics of the shallow and deep models. This contrasts with the Transformer experiments, where increasing the depth leads to an unstable training process.

\begin{figure}
\centering
\includegraphics[width=\linewidth]{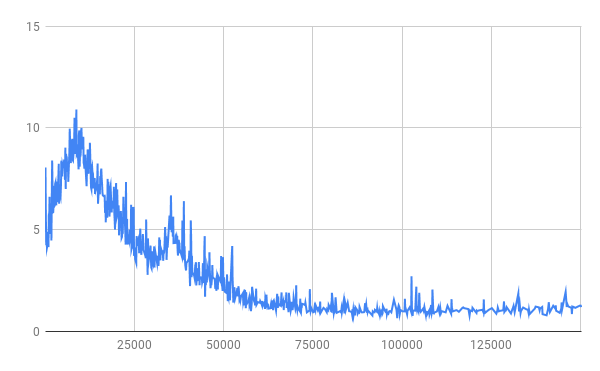}
\caption{Grad-norm ratio ($r_t$) vs training step for 20 layer Transformer with transparent attention.}
\label{fig:transparent_trans_grad_norm}
\end{figure}

To gain further insights into the stability of the two architectures we completely remove the residual connections from their encoders. Residual connections have been shown, in theory and practice, to improve training stability and performance of deeper networks (see \cite{ResNet,GradientsExplode,IdentityMatters,Eliminate}). Removing residual connections leads to disastrous results for the Transformer, where the training process either does not converge or results in significantly worse results. On the other hand, the 6 layer RNMT+ converges with only a slight degradation in quality. Deeper versions of RNMT+ fail to train in the absence of residual connections.

\begin{table*}[h]
\centering
\begin{tabular}{c|c|c|c|c|c}
\hline
\hline
En$\rightarrow$De WMT 14     & \multicolumn{4}{c|}{\begin{tabular}[c]{@{}l@{}}Transformer (Base)\end{tabular}} & \multicolumn{1}{c}{\begin{tabular}[c]{@{}l@{}}(Big)\end{tabular}} \\ \hline
Encoder layers  & 6 & 12 & 16 & 20 & 6\\ \hline
Num. Parameters  & 94M & 120M & 137M & 154M & 375M\\ \hline \hline
Baseline & 27.26 & * & * & * & 27.94 \\ \hline
Baseline - residuals & * & 6.00 & * & * & N/A\\ \hline
Transparent & 27.52 & 27.79 & \textbf{28.04} & 27.96 & N/A \\ \hline
\end{tabular}
\caption{\label{tab:En-De-trans}BLEU scores on En$\rightarrow$De newstest 2014 with Transformers. * indicates that a model failed to train.}
\end{table*}

\begin{table*}[h]
\centering
\begin{tabular}{c|c|c|c|c|c}
\hline
\hline
Cs$\rightarrow$En WMT 15     & \multicolumn{4}{c|}{\begin{tabular}[c]{@{}l@{}}Transformer (Base)\end{tabular}} & \multicolumn{1}{c}{\begin{tabular}[c]{@{}l@{}}(Big)\end{tabular}}\\ \hline
Encoder layers  & 6 & 12 & 16 & 20 & 6 \\ \hline
Num. Parameters  & 94M & 120M & 137M & 154M & 375M\\ \hline \hline
Baseline & 27.20 & * & * & * & 27.76 \\ \hline
Baseline - residuals & 25.83 & * & * & * & N/A \\ \hline
Transparent & 27.41 & 27.69 & \textbf{27.93} & 27.80 & N/A \\ \hline
\end{tabular}
\caption{\label{tab:Cs-En-trans} BLEU scores Cs$\rightarrow$En newstest 2015 with Transformers. * indicates that a model failed to train.}
\end{table*}

\begin{table*}[h]
\centering
\begin{tabular}{c|c|c|c|c|c}
\hline
\hline
En$\rightarrow$De WMT 14     & \multicolumn{4}{c|}{\begin{tabular}[c]{@{}l@{}}RNMT+ (512)\end{tabular}} & \multicolumn{1}{c}{\begin{tabular}[c]{@{}l@{}}(1024)\end{tabular}} \\ \hline
Encoder layers  & 6 & 12 & 16 & 20 & 6 \\ \hline
Num. Parameters  & 128M & 165M & 191M & 216M & 379M \\ \hline \hline
Baseline & 26.63 & 26.32 & 26.49 & 26.33 & \textbf{28.49} \\ \hline
Baseline - residuals & 26.37 & * & * & * & N/A \\ \hline
Transparent & 26.61 & 26.87 & 27.07 & \textbf{27.33} & N/A \\ \hline
\end{tabular}
\caption{\label{tab:En-De-rnmt}BLEU scores on En$\rightarrow$De newstest 2014 with RNMT+. * indicates that a model failed to train.}
\end{table*}

\begin{table*}[h]
\centering
\begin{tabular}{c|c|c|c|c|c}
\hline
\hline
Cs$\rightarrow$En WMT 15     & \multicolumn{4}{c|}{\begin{tabular}[c]{@{}l@{}}RNMT+ (512)\end{tabular}} & \multicolumn{1}{c}{\begin{tabular}[c]{@{}l@{}}(1024)\end{tabular}} \\ \hline
Encoder layers  & 6 & 12 & 16 & 20 & 6 \\ \hline
Num. Parameters  & 128M & 165M & 191M & 216M & 379M \\ \hline \hline
Baseline & 25.77 & 25.86 & 26.02 & 25.75 & 26.66 \\ \hline
Baseline - residuals & 25.43 & * & * & * & N/A \\ \hline
Transparent & 26.69 & 26.74 & \textbf{26.79} & 26.72 & N/A\\ \hline
\end{tabular}
\caption{\label{tab:Cs-En-rnmt} BLEU scores Cs$\rightarrow$En newstest 2015 with RNMT+. * indicates that a model failed to train.}
\end{table*}

\subsection{Regulating Deep Encoder Gradients with Transparent Attention}
Our baseline experiments reveal that mechanisms to regulate gradient flow can be critical to improving the optimization of deeper encoders. Since the only difference between our shallow and deep models is the number of layers in the encoder, the trainability issues are likely to be associated with gradient flow through the encoder.

\begin{figure}
\centering
\includegraphics[width=\linewidth]{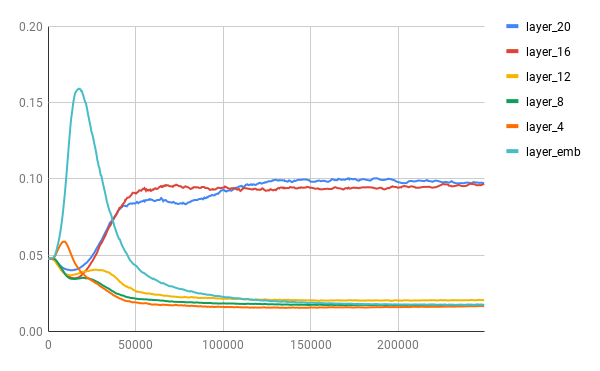}
\caption{Plot illustrating the variations in the learned attention weights $s_{i,6}$ for the 20 layer Transformer encoder over the training process.}
\label{fig:transformer_transparent_weights}
\end{figure}

To improve gradient flow we let the decoder attend weighted combinations of all encoder layer outputs, instead of just the top encoder layer. Similar approaches have been found to be useful in deep convolutional networks, for example \cite{WeightedResNet,DenseNet,Highway,StochasticDepth}, but this remains un-investigated in sequence-to-sequence models. We formulate our proposal below.

Assume the model has $N$ encoder layers and $M$ encoder-decoder attention modules. For Transformer models each decoder layer attends the encoder, so $M$ is equivalent to the number of decoder layers ($M=6$). For RNMT+, attention is only applied in the first decoder layer, thus $M=1$. 
Let the activations from the $i$-th encoder layer be $\{h^i_t| t=1\ldots T\}$, and embeddings be layer $0$. Then the traditional attention module attends to $\{h^N_t| \quad t = 1 \ldots T\}$. In transparent attention we evaluate $M$ weighted combinations of the encoder outputs, one corresponding to each attention module. We define a $(N+1) \times M$ weight vector $W$, which is learned during training.\footnote{Here +1 is for the embedding layer.} We apply dropout to $W$ since we empirically found it helpful to stabilize training. We then compute softmax $s$ to normalize the weights.
\begin{equation}
s_{i,j} = \frac{e^{W_{i, j}}}{\Sigma_{k=0}^{N} e^{W_{k, j}}} ,\quad j=1\ldots M
\end{equation}
We now define
\begin{equation}
z^j_t = \Sigma_{i=1}^{N+1} s_{i,j}h^i_t ,\quad t=1\ldots T ,\quad j=1\ldots M
\end{equation}
Now attention module $j$ attends to $\{z^j_t| \quad t=1\ldots T\}$.
Since in RNMT+ a projection is applied to the encoder final layer output, we apply a projection to the weighted combination of encoder outputs before the attention module.

\section{Results and Analysis}
Our results, from tables~\ref{tab:En-De-trans} and~\ref{tab:Cs-En-trans}, indicate that adding transparent attention improves the performance of most of our transformer experiments, but the gains are most pronounced for deeper models. While the baseline transformer fails to train with 12 layers or deeper encoders, transparent attention allows us to train encoders with up to 20 layers, improving by more than 0.7 BLEU points on both datasets. Relative to Transformer Big, deeper models seem to result in better or comparable performance with less than half the model capacity.

We also observe gains of 0.7 and 1.0 BLEU for RNMT+ models, on En$\rightarrow$De and Cs$\rightarrow$En respectively, as indicated by Tables~\ref{tab:En-De-rnmt} and~\ref{tab:Cs-En-rnmt}. However, experiments comparing wide models against deeper ones are inconclusive. While deeper models perform slightly better than a wide model with double their capacity on Cs-En, they are clearly out-performed by the larger model on En-De.

The $r_t$ plot in Figure~\ref{fig:transparent_trans_grad_norm}, also indicates that the learning dynamics now resemble what we expect to see with stable training. We also notice that the scale of $r_t$ now resembles that of the RNMT+ model, although the lower layers converge more slowly for the Transformer, possibly because it uses a much smaller learning rate.

A plot of the weights $s_{i,j}$, in Figure~\ref{fig:transformer_transparent_weights}, also seems to support our findings. The scalar weights for the lowest embeddings layer grow rapidly in the early stages of training, but once these layers converge the weights for layers 16 and 20 become much larger. The weights for the top few layers remain comparable at convergence, suggesting that the observed gains in performance might also be partially associated with an ensembling effect of the encoder features, similar to the effect observed in \cite{ELMO}.
\newpage
\section{Conclusions and Future Work}
In this work we explore deeper encoders for Transformer and RNMT+ based machine translation models. We observe that Transformer models are extremely difficult to train when encoder depth is increased beyond 12 layers. While RNMT+ models train with deeper encoders, we did not observe any big performance improvements. 

We associated the difficulty in training deeper encoders with hindered gradient flow, and resolved it by proposing the transparent attention mechanism. This enabled us to successfully train deeper Transformer and RNMT+ models, resulting in consistent gains in translation quality on both WMT'14 En$\rightarrow$De and WMT'15 Cs$\rightarrow$En.

Our results show that there is potential for improvement in translation quality by training deeper architectures, even though they pose optimization challenges. While this study explores training deeper encoders for narrow models, we plan to further study extremely deep and wide models to utilize the full strength of these architectures. 

\section{Acknowledgments}
We would like to thank the Google Brain and Google Translate teams for their foundational contributions to this project.

\bibliographystyle{acl_natbib_nourl}
\bibliography{em}



\end{document}